\documentclass{article}

\PassOptionsToPackage{numbers, compress}{natbib}
\usepackage[final]{nips_2017}

\usepackage[utf8]{inputenc} % allow utf-8 input
\usepackage[T1]{fontenc}    % use 8-bit T1 fonts
\usepackage{hyperref}       % hyperlinks
\usepackage{url}            % simple URL typesetting
\usepackage{booktabs}       % professional-quality tables
\usepackage{amsfonts}       % blackboard math symbols
\usepackage{nicefrac}       % compact symbols for 1/2, etc.
\usepackage{microtype}      % microtypography

\usepackage{hyperref}
\usepackage{url}
\usepackage{multirow}
\usepackage{graphicx}

\usepackage{amsmath}

\newcommand{\vm}{{\mathbf{m}}}
\newcommand{\vx}{{\mathbf{x}}}

\newcommand{\vz}{{\mathbf{z}}}

\newcommand{\vtheta}{{\boldsymbol{\theta}}}

\title{Semi-Supervised and Active Few-Shot Learning \\with Prototypical Networks}

\author{
  Rinu Boney\\
  The Curious AI Company \& Aalto University\\
  \texttt{rinu@cai.fi}
  \And
  Alexander Ilin\\
  The Curious AI Company\\
  \texttt{alexilin@cai.fi}
}

\begin{document}
% \nipsfinalcopy is no longer used

\maketitle

\begin{abstract}
We consider the problem of semi-supervised few-shot classification where a classifier needs to adapt to new tasks using a few labeled examples and (potentially many) unlabeled examples. We propose a clustering approach to the problem. The features extracted with Prototypical Networks \citep{snell2017prototypical} are clustered using $K$-means with the few labeled examples guiding the clustering process. We note that in many real-world applications the adaptation performance can be significantly improved by requesting the few labels through user feedback. We demonstrate good performance of the active adaptation strategy using image data.
\end{abstract}

\section{Introduction}

Few-shot learning addresses the problem of learning new concepts quickly, which is one of the important properties of human intelligence. In few-shot classification, the task is to adapt a classifier to previously unseen classes from just a few examples \citep{lake2015human,koch2015siamese,vinyals2016matching,ravi2016optimization}. This skill is useful in many practical applications since data annotation is laborious and a training set can be rather small. Recent approaches have focused on casting few-shot learning as a meta-learning problem, in which a model is trained on a variety of tasks to adapt quickly using a small number of training examples \citep{finn2017model, munkhdalai2017meta, ravi2016optimization, li2017meta}. Thus, the system learns to adapt to new problems using few labeled samples based on the knowledge transferred from its previous experience with related problems.

In some real-world problems, the tasks to which the learning system needs to adapt may contain both (few) labeled and (many) unlabeled examples, the problem known as semi-supervised learning. For example, a common feature of photo management applications is the automatic organization of images based on limited interactive supervision from the user.
The classes relevant to a specific user are likely to be different from the classes in publicly available image datasets such as ImageNet, thus there is a need for adaptation. The user can facilitate the adaptation by labeling a few images by personal preferences. In this application, the learning system also has access to lots of unlabeled images and it can make use of those to improve the classification accuracy. This is the problem of semi-supervised few-shot classification that we consider.

In this paper, we show how to tackle this problem using a model called Prototypical Networks (PN) \citep{snell2017prototypical}.
Using the observation that PN tends to produce clustered data representations, we cast the semi-supervised
few-shot problem as a semi-supervised \emph{clustering} problem and propose several approaches to solve it.
We also take inspiration from \citep{cohn2003semi} to enable adaptation to new classification tasks using feedback from the user. We argue that this approach can be practical in many real-world applications, as
many use cases of semi-supervised few-shot adaptation imply interaction with a user and therefore active
learning is often possible.

We use the following formulation of the few-shot classification problem. There is a training set which consists of a large set of classes and we have access to (potentially many) labeled samples from each class in the training set. At test time, the task is to separate $N$ previously unseen classes
using a small number $k$ of labeled samples and (potentially many) unlabeled samples from the same $N$ classes.
This is called an $N$-way $k$-shot classification task.
We follow the recent literature and use the episodic regime of training and evaluation, as we explain in the following section.

\section{Prototypical Networks}
\label{sc:pn}

The episodic training of PN iterates between the following steps.
A subset of $N$ classes is randomly selected to formulate one training task. 
For each training task, a support set $S=\{(\vx_1, y_1), ..., (\vx_{n}, y_{n})\}$
and a query set $Q=\{(\vx_{n+1},y_{n+1}), ..., (\vx_{n+m}, y_{n+m})\}$ are created by sampling examples from the selected classes,
where $\vx_j$ are inputs and $y_j$ are the corresponding labels.
% TODO: The notation for S and Q is not optimal.

Prototypical Networks compute representations of the inputs $\vx$ using an embedding function $g$ parameterized with $\vtheta$: $\vz = g(\vx, \vtheta)$. Each class $c$ is represented in the embedding space by a prototype vector which is computed as the mean vector of the embedded inputs for all the examples $S_c$ of the corresponding class $c$:
\begin{align}
  \vm_c &= \frac{1}{|S_c|} \sum_{(\vx_j, y_j)\in S_c} g(\vx_j, \vtheta) %\vz_j
\,.
\label{eq:z_k}
\end{align}

The distribution over predicted labels $y$ for a new sample is computed using softmax over negative distances to the prototypes in the embedding space:
\begin{align}
  p(y = c| \vx, \{\vm_c\}) = \frac{\exp(-d(\vz, \vm_c))}{\sum_{c'} \exp(-d(\vz, \vm_{c'}))}
\,.
\label{eq:pn_p}
\end{align}

Parameters $\vtheta$ are updated so as to improve the likelihood computed on the query set:
\[
\sum_{(\vx_j,y_j)\in Q} \log p(y=y_j | \vx_j, \{\vm_c\} )
\,,
\]
which is computed using \eqref{eq:pn_p} with the estimated prototypes.

\section{Adaptation Using Unlabeled Data}
\subsection{Semi-Supervised Few-Shot Adaptation}
\label{sec:few_shot}

In the semi-supervised scenario, we need to adapt to tasks which contain both labeled and unlabeled samples.
Classical methods of semi-supervised learning \citep{chapelle2006semi} are based on making certain assumptions such
as the smoothness assumption (the label function is smoother in high-density data regions than in low-density regions),
the cluster assumption (points in the same cluster are likely to be in the same class), the manifold assumption (the data lie
on a low-dimensional manifold).
Many recent algorithms in the deep learning literature are built around the smoothness and the manifold assumptions
\citep[see, e.g.,][]{rasmus2015semi,miyato2015distributional,laine2016temporal,tarvainen2017weight} which are
induced by extra terms in the objective function which depends on unlabeled data.

In this paper, we address semi-supervised classification using the cluster assumption. The motivation for this comes
from the observation that PN tends to produce clustered data representations in the embedding space. This is induced by the PN
decision rule \eqref{eq:pn_p} which uses the distances of a sample to the class means. The clustering approach to semi-supervised
learning is often referred as semi-supervised clustering \citep{basu2002semi}.

Our proposed algorithm works as follows. PN is trained with a standard training procedure which involves sampling of tasks, computing
the prototypes of each class and updating parameters $\vtheta$ of the embedding network using stochastic gradient descent, where
the gradient is computed using samples of the query set.
At test time, the prototypes are first estimated with the labeled data using \eqref{eq:z_k}.
We then perform the standard $K$-means algorithm \citep{lloyd1982least} on the embeddings of both labeled and unlabeled data initializing
the cluster means with the prototypes computed from the labeled data.
This corresponds to the seeding approach of $K$-means proposed by \citet{basu2002semi}.
The algorithm typically converges in just a few iterations (see Table~\ref{t:miniimagenet_semi_iter}).
We have also tried the constrained $K$-means approach in which the cluster membership of the labeled examples is never changed.
Both approaches yielded similar results and therefore we present only the results of the seeding approach which was slightly better.

The proposed algorithm is related to the one concurrently developed by \citet{ren1meta} who do semi-supervised few-shot classification
using the constrained $K$-means clustering. The main difference is that they perform clustering also at training time and therefore they
use soft cluster assignments to keep the computational graph differentiable. They do only one iteration of $K$-means, as doing more iterations
does not improve the performance. We obtained similar results using soft cluster assignment and found experimentally that $K$-means with hard clustering behaves more robustly (see results in Table~\ref{t:miniimagenet_semi_iter}).
Furthermore, \citet{ren1meta} consider the scenario in which the unlabeled support set may contain samples from irrelevant classes.

\begin{figure}[t]
\begin{center}
\includegraphics[width=0.32\textwidth, trim={25 15 30 25},clip]{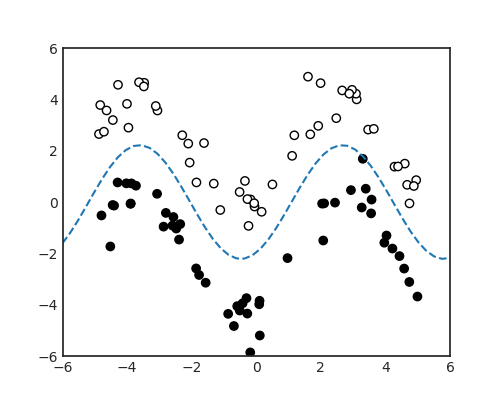}
\includegraphics[width=0.32\textwidth, trim={25 15 30 25},clip]{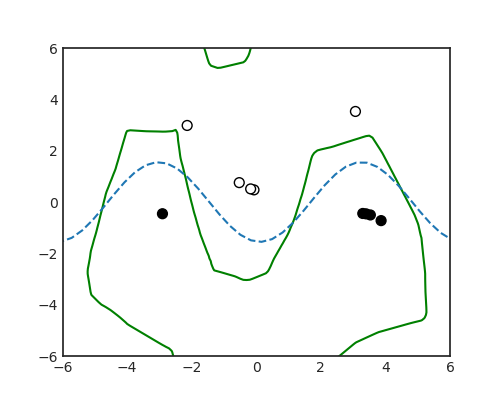}
\includegraphics[width=0.32\textwidth, trim={25 15 30 25},clip]{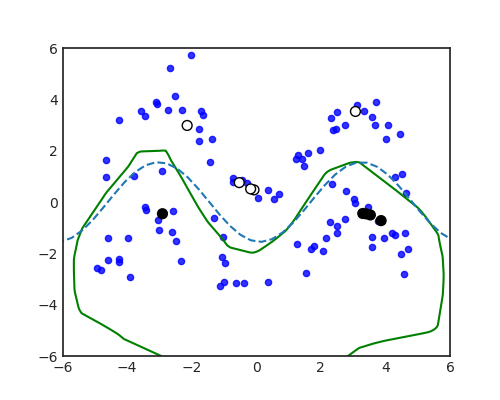}
\end{center}
\caption{Left: Example task from the sine data set. The black dots correspond to samples of one class and
the white dots correspond to samples from the other class. The optimal decision boundary is shown with the dashed blue line. 
Middle: Example of adaptation to a test task from 10 labeled samples.
Right:  Example of adaptation to a test task from the same 10 labeled samples and 100 unlabeled samples.
The blue dots correspond to unlabeled samples.}
\label{f:sine}
\end{figure}

\subsection{Unsupervised Adaptation}

In many practical applications, the classification tasks to which a classifier needs to adapt share the same
output space. As an example,
consider a material recognition system deployed in multiple factories. The same categories of materials
need to be recognized at different sites, however, factories may have slightly different lighting conditions
or other factors affecting the recognition process,
motivating the need for adaptation of each recognition system. In these scenarios, since the classes do not change across tasks,
the average class prototypes can be computed at the end of training by averaging the prototypes of the corresponding
class from all the training tasks.
At test time, we cluster the samples using $K$-means and then assign each cluster to the class of the closest prototype.

\subsection{Experiments with Synthetic Data}
\label{sec:sine}

To test the proposed algorithm, we created a synthetic data set which is fast to experiment with.
The dataset consists of a set of two-dimensional classification tasks with two classes,
in which the optimal decision boundary is a sine wave  (see Fig.~\ref{f:sine}). The amplitude $A$
of the optimal decision boundary varies across tasks within $[0.1, 5.0]$ and the phase $\phi$ varies within $[0, \pi]$.
The first dimension of the data samples is drawn uniformly from $[-5, 5]$ and the second dimension is computed as
\[
x_2 = A \sin(x_1 + \phi) + \epsilon
\]
where $\epsilon$ is a noise term with the Laplace distribution with the mean $\pm2$ (depending on the class) and
the scale parameter 0.5. We sampled 100 tasks for training and 1000 tasks for testing. 

In this experiment, we use a fully connected network with two hidden layers of size 40 with ReLU nonlinearity as the embedding network. Examples of the decision boundaries produced by PN on test tasks are shown in Fig.~\ref{f:sine}. Note
that even for a small number of training examples the PN decision boundaries resemble the sine wave,
thus the knowledge is transferred between tasks.
The classification errors of semi-supervised adaptation on test tasks depending on the number of unlabeled samples are shown in Table~\ref{t:sine}. One can see from the results that the PN accuracy improves with the use of unlabeled data.
However, the improvement plateaus and the results do not improve much after adding more than 100 data points.

We also experimented with the proposed unsupervised adaptation method on the synthetic dataset.
The classification errors as a function of the number of unlabeled samples are shown in Table~\ref{t:sine-active}.

\begin{table}[t]

\begin{minipage}[t]{0.49\linewidth}
\caption{Semi-supervised adaptation on sine data set: Classification errors obtained with 10 labeled samples and $n$ extra samples
(either labeled or unlabeled).}
\label{t:sine}
\begin{center}
\begin{tabular}{r|rr}
\toprule
$n$ & labeled & unlabeled \\
\midrule
0             &    6.38 &  6.38 \\
10            &    4.97 &  5.54 \\
100           &    2.73 &  4.70 \\
1000          &    1.68 &  4.57 \\
\bottomrule
\end{tabular}
\end{center}
%\end{table}
\end{minipage}
\hfill
%\begin{table}[h]
\begin{minipage}[t]{0.48\linewidth}
\caption{Fully-supervised and unsupervised adaptation on sine data set: Classification errors obtained with $n$ labeled
 or unlabeled samples.}
\label{t:sine-active}
\begin{center}
\begin{tabular}{r|rr}
\toprule
$n$ & supervised & unsupervised \\
\midrule
10             &  6.38 &  7.71 \\
100            &  2.98 &  6.40 \\
1000           &  1.99 &  6.39 \\
\bottomrule
\end{tabular}
\end{center}
\end{minipage}

\end{table}

\subsection{Experiments with miniImagenet}

We tested the proposed method in the semi-supervised scenario on the miniImagenet recognition task proposed by \citet{vinyals2016matching}. The dataset involves downsampled 84x84 images from 64 training classes, 12 validation classes, and 24 test classes from ImageNet. We use the same split as \cite{ravi2016optimization} and follow the experimental setup which involves $N$-way $k$-shot classification, similarly to \cite{vinyals2016matching} that is every task at test time contains $N$ classes with $k$ labeled examples from each class.
For the semi-supervised case, we also assume the existence of $M$ unlabeled samples per class at test time. Following previous works, we use 15 test samples per class for each task during training and for evaluation. We evaluate the model over 2400 tasks which involve the 24 classes reserved for testing.

One challenge with the miniImagenet data set is that it requires rather complex features but it contains a relatively
little amount of data. Therefore, preventing overfitting becomes an important issue. Most previous works
\cite{vinyals2016matching,ravi2016optimization,finn2017one,snell2017prototypical} used conventional convolutional networks with a small number of layers to prevent overfitting. For comparability, we use the same architecture which consists of four blocks. Each block consists of a $3 \times 3$ convolutional layer with 64 channels followed by batch normalization, ReLU non-linearity and a $2 \times 2$ max-pooling layer which results in an embedding space of dimensionality 1600. In Prototypical Networks \cite{snell2017prototypical}, the model was fine-tuned by having more classes at training time than at test time (e.g., 30-way training and 5-way testing) and a learning rate decay schedule. We use a constant learning rate and 5-way training for simplicity. This simple network may not be expressive enough to capture relevant complex features. So, we also consider a Wide Residual Network \cite{zagoruyko2016wide} as the embedding network. We use a network of depth 16 and a widening factor of 6. We also use a $8\times8$ pooling with a stride of 4  at the end to obtain embeddings of dimensionality 384. The network is regularized with dropout with a rate of $0.3$. We use the Adam optimizer \cite{kingma2014adam} with a learning rate of 0.01 for training the ResNet and 0.001 for training the four-block architecture. We perform early stopping to prevent overfitting.\footnote{Residual Networks are typically trained using stochastic gradient descent with momentum and we expect better results by doing the same and fine-tuning the hyperparameters.} We train the ResNet model with $N=30$ classes in each task for 1-shot classification and $N=20$ classes in each task for 5-shot classification, similarly to the original PN paper. For the ResNet model, we tried varying the number of classes during training and observed that it had a much smaller impact on the results compared to the observations in \cite{snell2017prototypical}. This suggests that tuning this regularization parameter is less important for this architecture.

The results presented in Table~\ref{t:miniimagenet} show that our approach scales to both feature extractor architectures. One can observe that using the Wide ResNets to learn the
embedding space yields noticeable improvements in the classification accuracy compared to the baseline methods.
The improvements are more significant for the 20-way classification. Table~\ref{t:miniimagenet} presents the results of the PN
tested in the semi-supervised scenario: The prototypes are re-estimated at test time using both labeled samples and the inputs from the query set.

In Table~\ref{t:miniimagenet_semi_nsamp}, we show how the number of unlabeled examples at test time affects the classification accuracy of the trained PN. The results indicate that more unlabeled samples yield better performance, however, the improvement plateaus very quickly with the increase of the number of unlabeled samples. This agrees with the results obtained on the synthetic data reported in Section~\ref{sec:sine}. Notably, the classification performance of the four-block architecture scales well with increasing the number of unlabeled samples, closely matching the performance of the ResNet in the case of 120 unlabeled samples per class.
The evolution of the classification accuracy with increasing the number of $K$-means iterations is shown in Table~\ref{t:miniimagenet_semi_iter}.

\begin{table}[t]
\caption{Average classification accuracy (with 95\% confidence intervals) on miniImagenet.
The comparison numbers for the 20-way testing are from \cite{li2017meta}.}
\label{t:miniimagenet}
\begin{center}
\begin{tabular}{l|ll|ll}
\toprule
                                & \multicolumn{2}{|c|}{5-way testing} & \multicolumn{2}{|c}{20-way testing}\\
\midrule
Model                           & 1-shot           & 5-shot & 1-shot           & 5-shot\\
\midrule
fine-tuning baseline             & $28.86 \pm 0.54$ & $49.79 \pm 0.79$ & -- & -- \\
nearest-neighbor baseline        & $41.08 \pm 0.70$ & $51.04 \pm 0.65$ & -- & -- \\
Meta-LSTM \cite{ravi2016optimization}    & $43.44 \pm 0.77$ & $60.60 \pm 0.71$ & $16.70 \pm 0.23$ & $26.06 \pm 0.25$ \\
Matching nets \cite{vinyals2016matching} & $46.6$ & $60.0$ & $17.31 \pm 0.22$ & $22.69 \pm 0.20$ \\
MAML \cite{finn2017one}         & $48.70 \pm 1.84$ & $63.11 \pm 0.92$ & $16.49 \pm 0.58$ & $19.29 \pm 0.29$ \\
ARC \cite{shyam2017attentive}   & $49.14$ & -- & -- & -- \\
Meta Networks \cite{munkhdalai2017meta}       & $49.21 \pm 0.96$ & -- & -- & -- \\
PN \cite{snell2017prototypical} & $49.42 \pm 0.78$ & $68.20 \pm 0.66$  & -- & -- \\
Meta-SGD \cite{li2017meta}               & $50.47 \pm 1.87$ & $64.03 \pm 0.94$ & $17.56 \pm 0.64$ & $28.92 \pm 0.35$ \\
PN (ours)                 & $48.06 \pm 0.82$ & $64.65 \pm 0.72$ & $20.36 \pm 0.24$ & $34.42 \pm 0.23$ \\
PN, re-estimate (ours)               & $51.07 \pm 0.90$ & $65.52 \pm 0.71$ & $21.06 \pm 0.26$ & $34.45 \pm 0.22$ \\
Resnet PN (ours)                 & $\pmb{51.69 \pm 0.42}$    & $\pmb{69.57 \pm 0.64}$ & $\pmb{23.35 \pm 0.28}$ & $\pmb{41.10 \pm 0.25}$ \\
Resnet PN, re-estimate (ours)   & $\pmb{54.05 \pm 0.47}$  & $\pmb{70.92 \pm 0.66}$ & $\pmb{23.59 \pm 0.31}$ & $\pmb{41.23 \pm 0.26}$ \\
\bottomrule
\end{tabular}
\end{center}
\end{table}

\begin{table}[t]
\caption{Average classification accuracy (with 95\% confidence intervals) on miniImagenet for the semi-supervised scenario for different number of unlabeled samples per class ($M$) available at test time.}
\label{t:miniimagenet_semi_nsamp}
\begin{center}
\begin{tabular}{ll|ll}
\toprule
 & & \multicolumn{2}{c}{5 way testing}\\
 & $M$ & 1-shot & 5-shot \\
\midrule
\multirow{4}{*}{PN (ours)} & 15 & $51.07 \pm 0.90$ & $65.52 \pm 0.71$ \\
& 30 & $53.41 \pm 0.94$ & $65.80 \pm 0.67$ \\
& 60 & $54.48 \pm 0.98$ & $65.94 \pm 0.69$ \\
& 120 & $55.23 \pm 1.04 $ & $65.96 \pm 0.68$ \\
\midrule
\multirow{4}{*}{Resnet PN} & 15 & $54.05 \pm 0.47$ & $70.92 \pm 0.66$ \\
& 30 & $54.70 \pm 0.46$ & $71.86 \pm 0.59$ \\
& 60 & $55.66 \pm 0.46$ & $72.21 \pm 0.55$ \\
& 120 & $55.67 \pm 0.45$ & $72.55 \pm 0.52$ \\
\bottomrule
\end{tabular}
\end{center}
\end{table}

\begin{table}[h]
\caption{Average classification accuracy (with 95\% confidence intervals) on miniImagenet for the semi-supervised scenario
as a function of the number of $K$-means iterations. Each task consists of 1 labeled sample and 15 unlabeled samples.}
\label{t:miniimagenet_semi_iter}
\begin{center}
\begin{tabular}{ll|ll}
\toprule
 &  & Hard $K$-means & Soft $K$-means \\
\midrule
\multirow{4}{*}{PN (ours)} & 0 & $48.06 \pm 0.82$ & $48.06 \pm 0.41$ \\
                    & 1 & $50.13 \pm 0.88$ & $51.36 \pm 0.75$ \\
                    & 2 & $50.76 \pm 0.93$ & $46.53 \pm 0.68$ \\
                    & 10 & $51.07 \pm 0.90$ & $37.16 \pm 0.34$ \\
\midrule
\multirow{4}{*}{Resnet PN} & 0 & $51.69 \pm 0.42$ & $51.69 \pm 0.42$ \\
                           & 1 & $53.42 \pm 0.48$ & $54.65 \pm 0.45$ \\
                           & 2 & $53.98 \pm 0.50$ & $50.62 \pm 0.41$ \\
                           & 10 & $54.05 \pm 0.47$ & $38.19 \pm 0.42$ \\
\bottomrule
\end{tabular}
\end{center}
\end{table}

\section{Active Few-Shot Adaptation}

There are two sources of errors which the semi-supervised adaptation algorithm proposed in Section~\ref{sec:few_shot}
can accumulate: 1)~errors due to incorrect clustering of data, 2)~errors due to incorrect labeling of the clusters.
The second type of errors can occur when the few labeled examples are outliers which end up closer to the prototype
of another class in the embedding space. In this paper, we advocate that the most practical way to correct the second type
of errors can be through user feedback, since in many applications of the semi-supervised few-shot adaptation,
interaction with the user is possible. This idea is inspired by the work of \citet{cohn2003semi} who introduced a clustering
approach that allows a user to iteratively provide feedback to a clustering algorithm.

Consider the previously introduced example of few-shot learning in photo management applications. Although it is possible to
ask the user to label a few photographs and use those labels to classify the rest of the pictures, it is 
extremely difficult and tiresome for the user to scroll through all the photos and decide which samples should be labeled.
Instead, using the observation that ``It is easier to criticize than to create'' \citep{cohn2003semi}, one can
initially cluster the photos and then request the user to label certain photos (or provide other types of feedback)
so that the data are properly clustered and labeled.
The user can provide feedback in various forms and therefore can effectively introduce various constraints that can further
guide the clustering process. For example, a user can assign the whole cluster to a particular class, assign a sample to
a particular cluster, mark that a particular sample does not belong to the assigned cluster, split and combine clusters. These constraints could be easily induced in basic clustering algorithms such as $K$-means.
For examples, \citet{wagstaff2001constrained} introduced constraints between samples in the data set such as must-link
(two samples have to be in the same cluster) and cannot-link (two samples have to be in different clusters) and the 
clustering algorithm finds a solution that satisfies all the constraints.

Even outside the context of few-shot learning, this active learning approach can be used to adapt a pre-trained classifier. Assume that we have a classifier that clusters the classes of a particular classification task such as ImageNet. Then, during test time it is possible to interactively split clusters to make coarse-grained classifications or to assign multiple clusters to a super-cluster (to make hierarchical predictions).

In this paper, we assume that the user can provide feedback only in the form of labeling a particular sample or
labeling the whole cluster.
We propose to use PN as a feature extractor, cluster the samples in the embedding space using $K$-means
and then label the clusters by requesting one labeled example for each cluster from the user.
For each cluster $c'$, we choose sample $\vz_{c'}$ to be labeled by the user by maximizing an acquisition function $a(\vz, c')$:
\[
  \vz_{c'} = \max_{\vz \in U_{c'}} a(\vz, c')
\,,
\]
where $U_{c'}$ is the set of embedded inputs belonging to cluster $c'$.
We explore a few acquisition functions:
\begin{itemize}
\item
\textbf{Random}: Sample a data point uniformly at random from each cluster. This is a baseline approach.

\item
\textbf{Nearest}: Select the data point which is closest to the cluster center:
\[
  a(\vz, c') = -d(\vz, \mathbf{m}_{c'})
\,,
\]
where $\mathbf{m}_{c'}$ is the mean (cluster center) of cluster $c'$.

\item
\textbf{Entropy}: Select the sample with the least entropy:
\[
  a(\vz, c') = \sum_{c} p(y=c|\vz) \log p(y=c|\vz)
\]
Thus, we select a sample with the least uncertainty that it belongs to a certain cluster.

\item
\textbf{Margin}: Select a sample with the largest margin between the most likely and second most likely labels.
\[
  a(\vz, c') = p(y=c_1(\vz)| \vz) - p(y=c_2(\vz)|\vz)
\]
where $c_1(\vz)$ and $c_2(\vz)$ are the most likely and the second most likely clusters of embedded input $\vz$ respectively.
This quantity was proposed as a measure of uncertainty by \citet{scheffer2001active}.

\end{itemize}

We also try to simulate a case when the user can label the whole cluster, as in some applications it can certainly be possible.
This approach directly measures the clustering accuracy and we call it ``oracle''.
\begin{itemize}
\item
\textbf{Oracle}: 
We label each cluster based on the distance of the cluster mean to the prototypes computed from the true labels of all the samples.
\end{itemize}

\subsection{Experiments with miniImagenet}

\begin{figure}
\centering
\begin{tabular}{cccccc}
 & True Labels & Optimal & Supervised & Semi-supervised & Active \\

(a) &
\raisebox{-.5\height}{\includegraphics[width=0.185\textwidth, trim={35 25 28 34},clip]{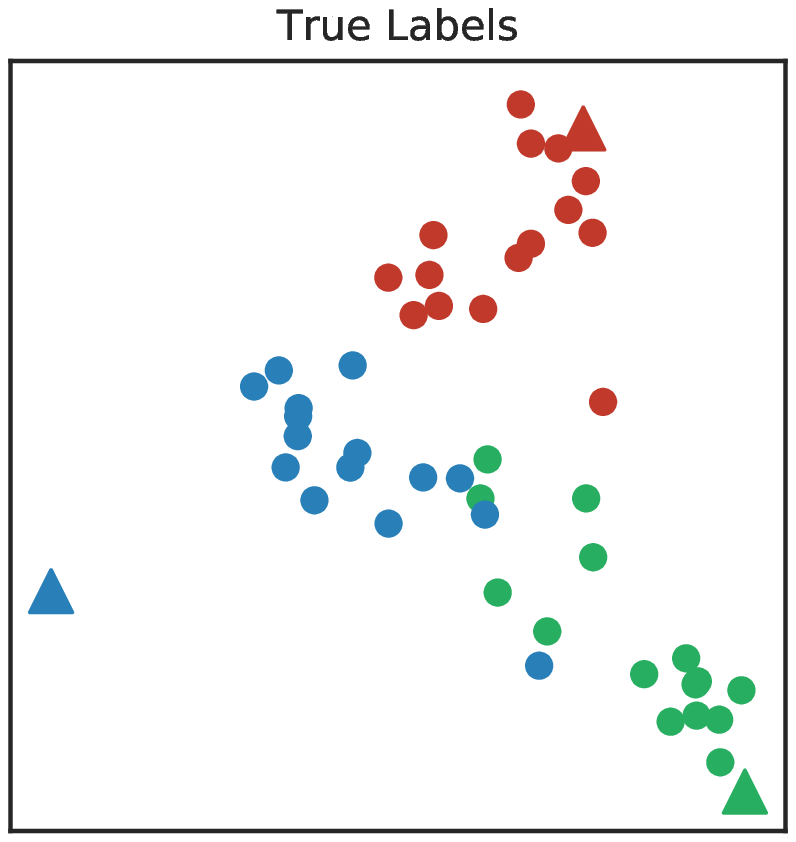}} &
\raisebox{-.5\height}{\includegraphics[width=0.185\textwidth, trim={35 25 28 34},clip]{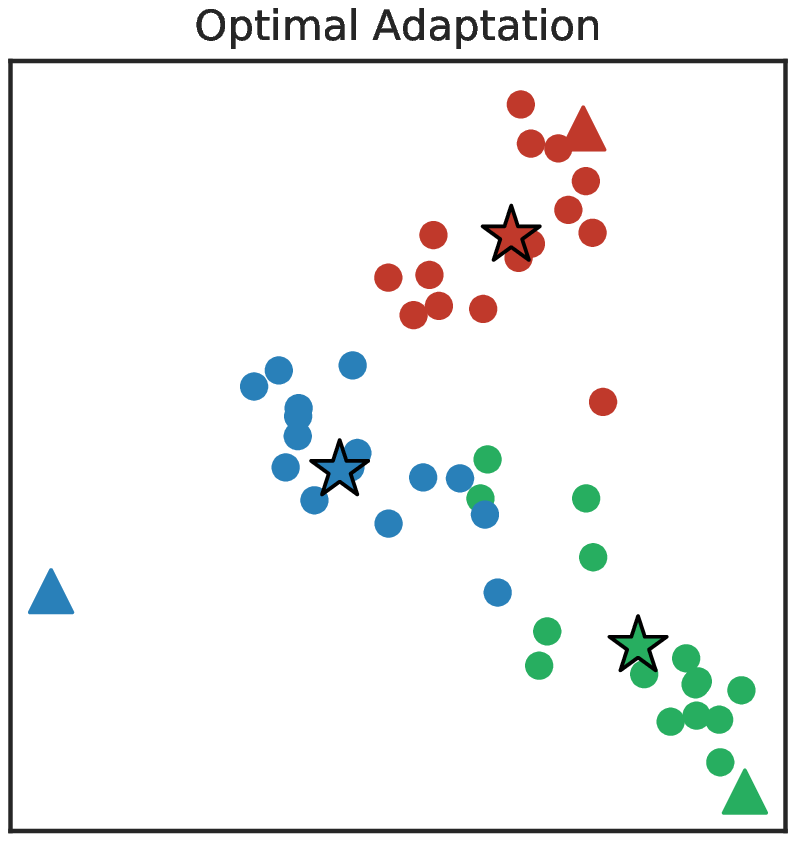}} &
\raisebox{-.5\height}{\includegraphics[width=0.185\textwidth, trim={35 25 28 34},clip]{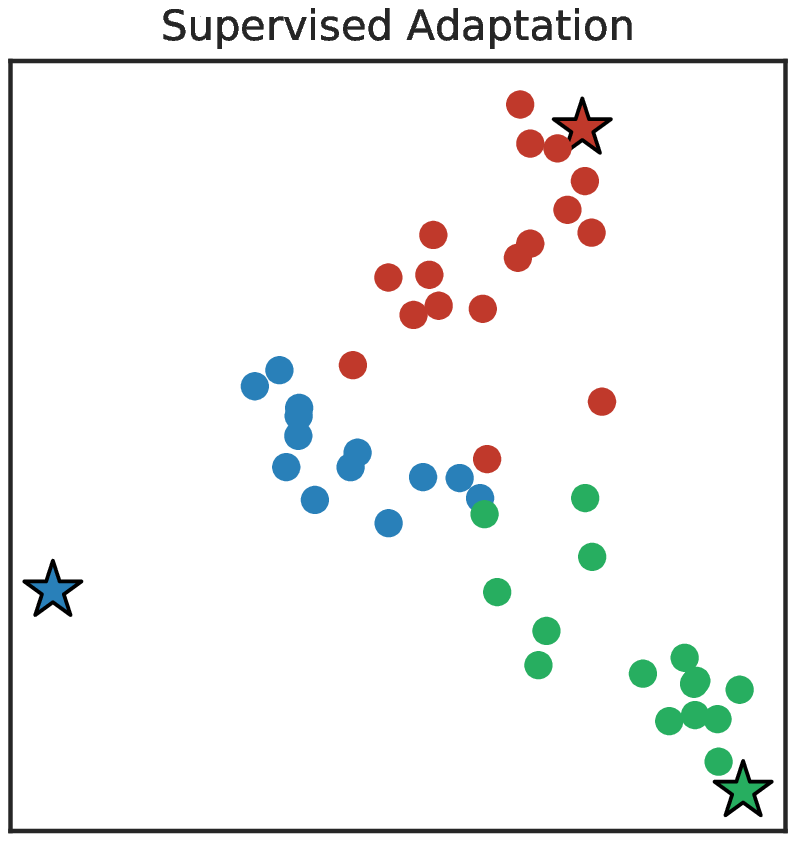}} &
\raisebox{-.5\height}{\includegraphics[width=0.185\textwidth, trim={35 25 28 34},clip]{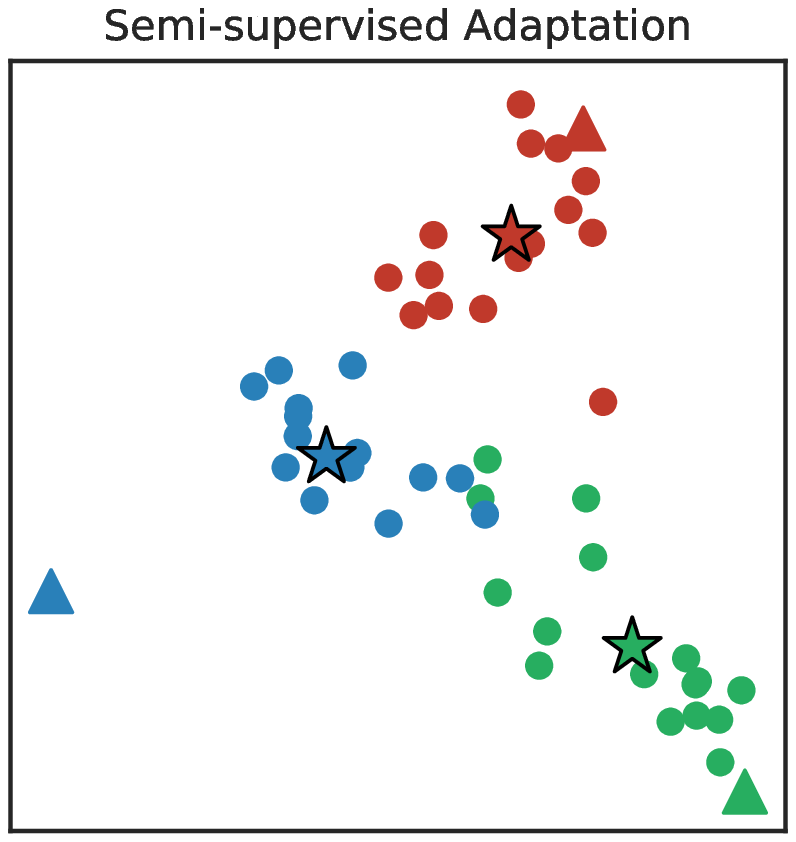}} &
\raisebox{-.5\height}{\includegraphics[width=0.185\textwidth, trim={35 25 28 34},clip]{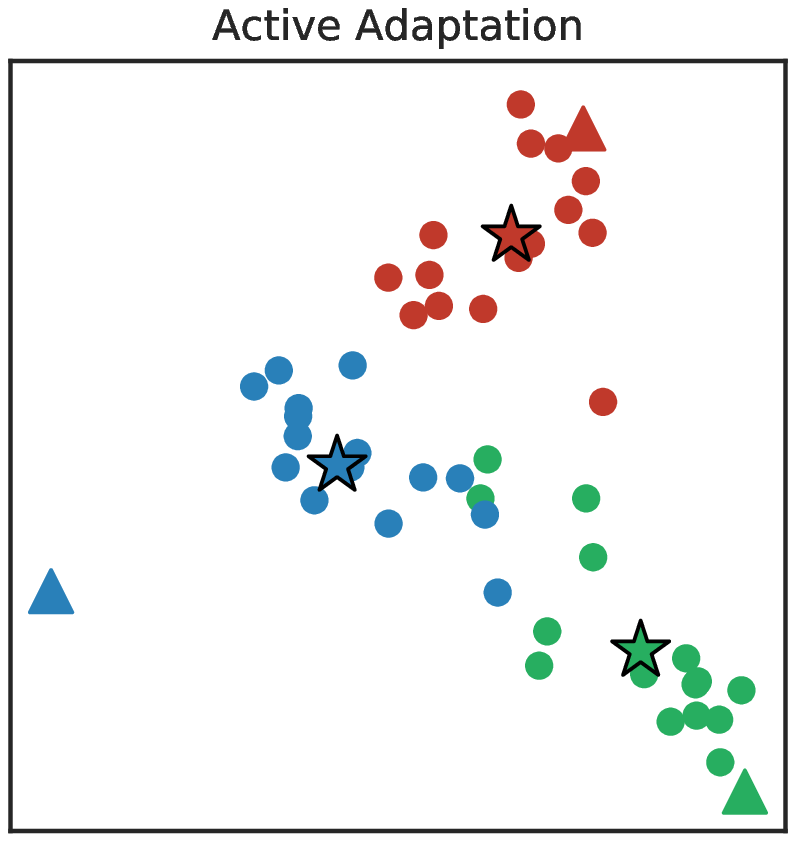}} \\[1.3cm]

(b) &
\raisebox{-.5\height}{\includegraphics[width=0.185\textwidth, trim={35 25 28 34},clip]{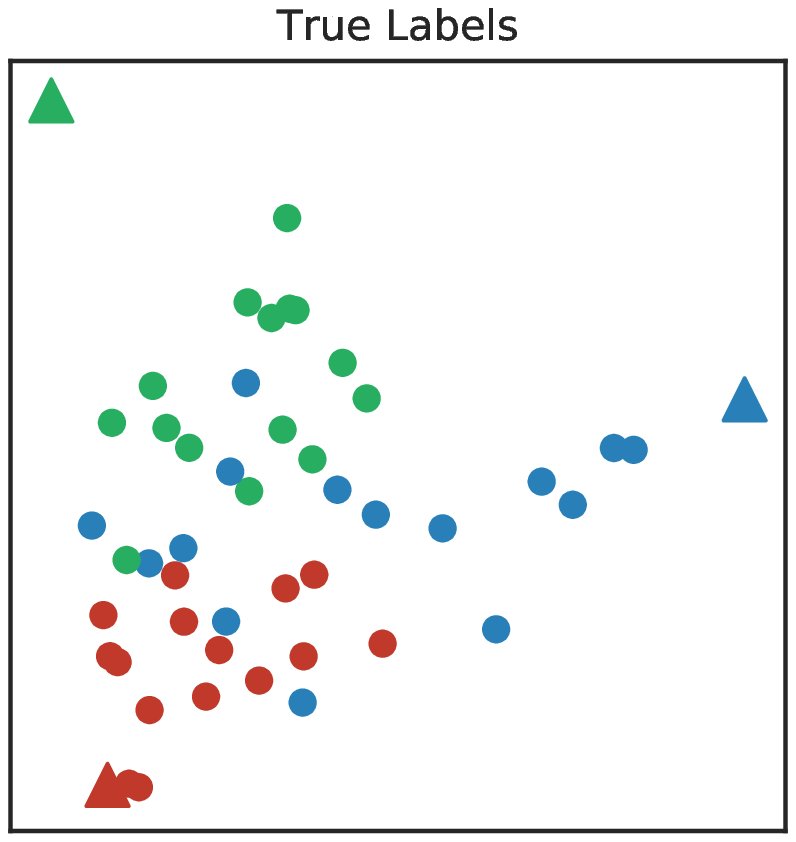}} &
\raisebox{-.5\height}{\includegraphics[width=0.185\textwidth, trim={35 25 28 34},clip]{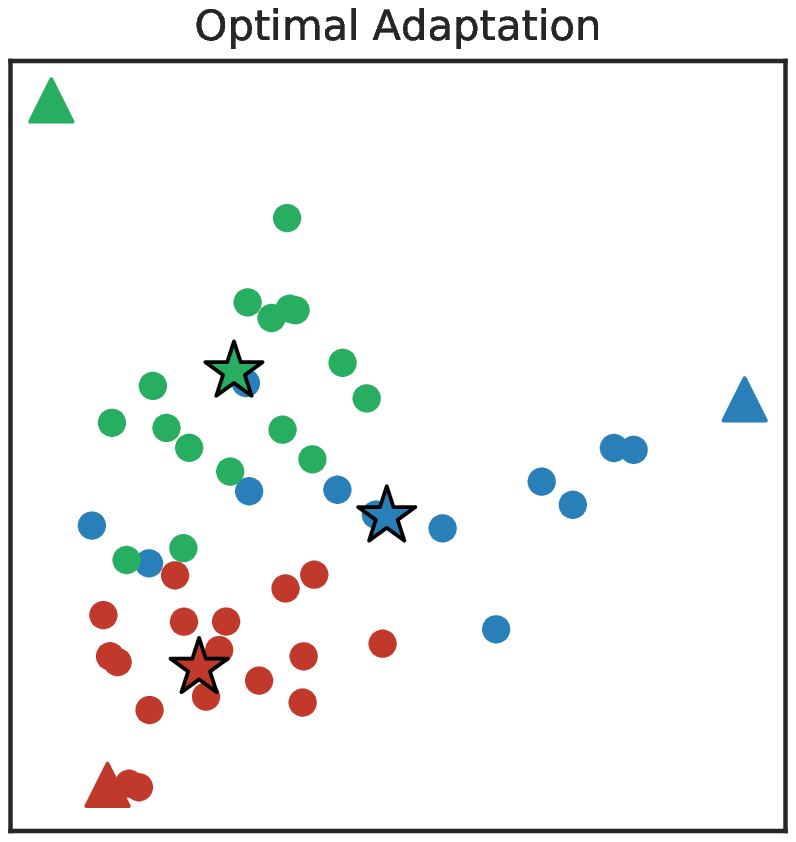}} &
\raisebox{-.5\height}{\includegraphics[width=0.185\textwidth, trim={35 25 28 34},clip]{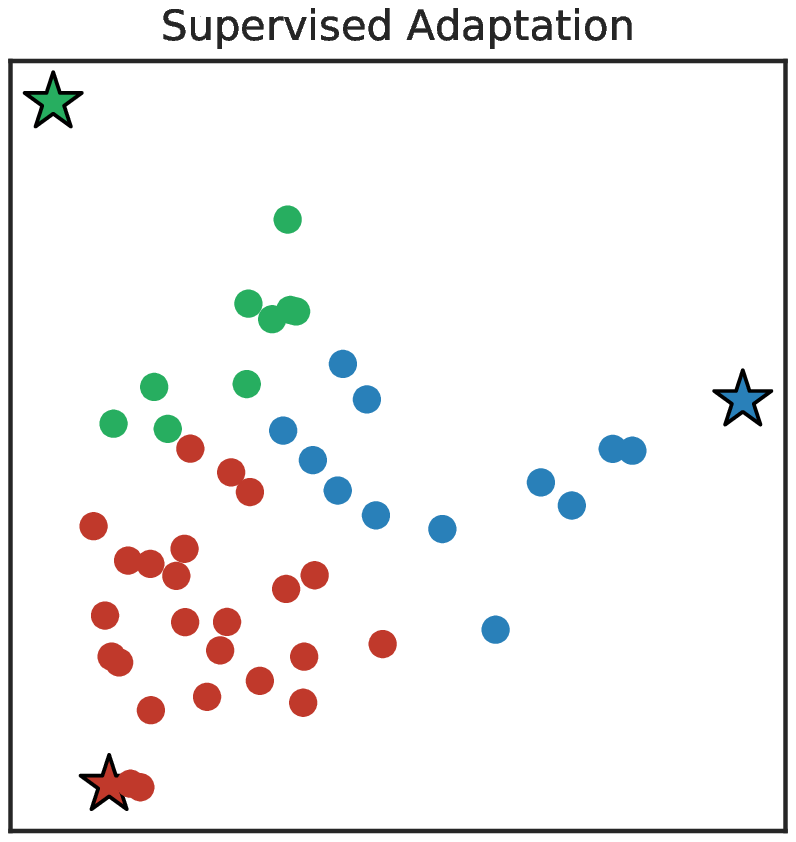}} &
\raisebox{-.5\height}{\includegraphics[width=0.185\textwidth, trim={35 25 28 34},clip]{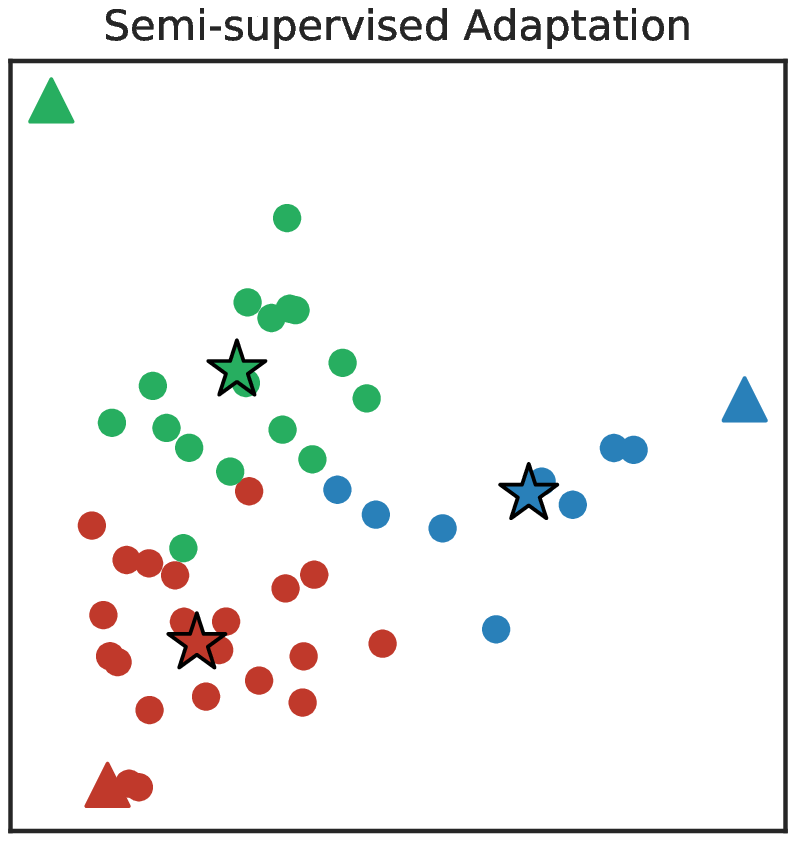}} &
\raisebox{-.5\height}{\includegraphics[width=0.185\textwidth, trim={35 25 28 34},clip]{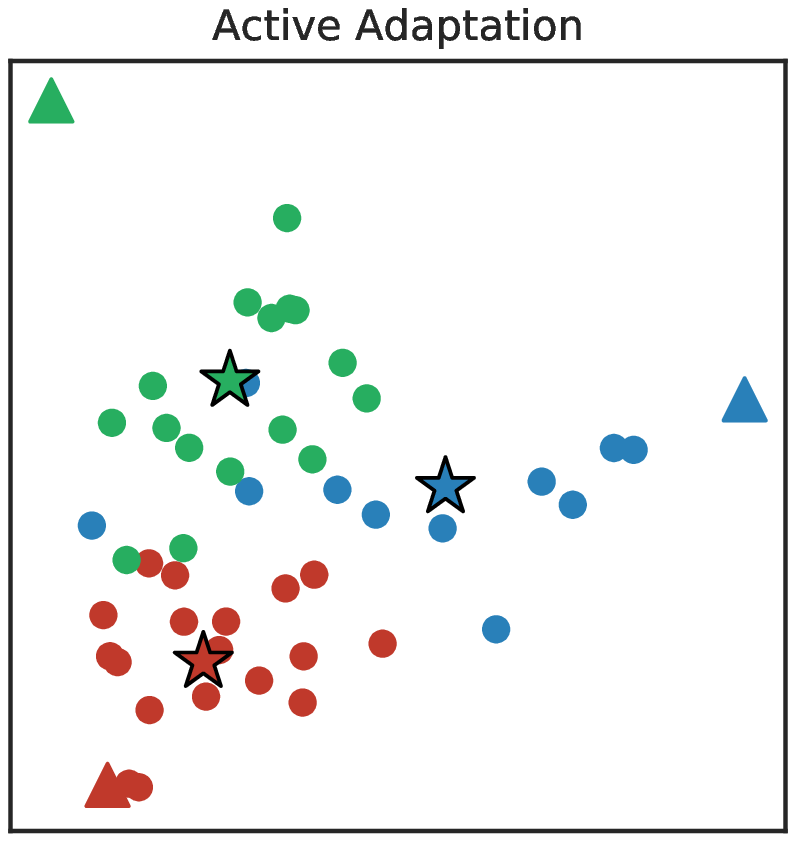}} \\[1.3cm]

(c) &
\raisebox{-.5\height}{\includegraphics[width=0.185\textwidth, trim={35 25 28 34},clip]{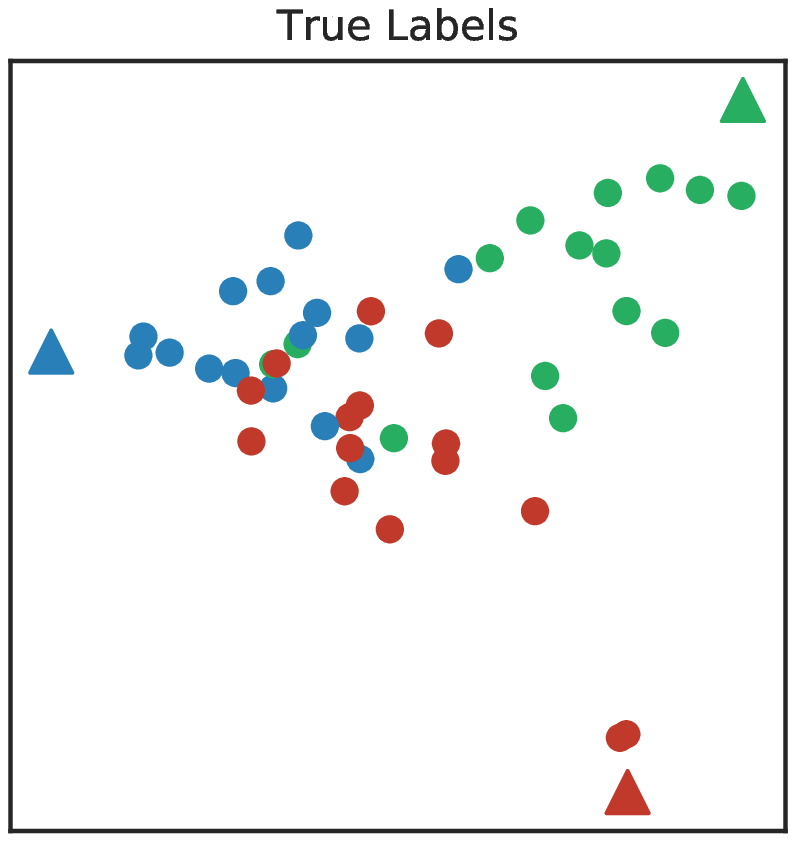}} &
\raisebox{-.5\height}{\includegraphics[width=0.185\textwidth, trim={35 25 28 34},clip]{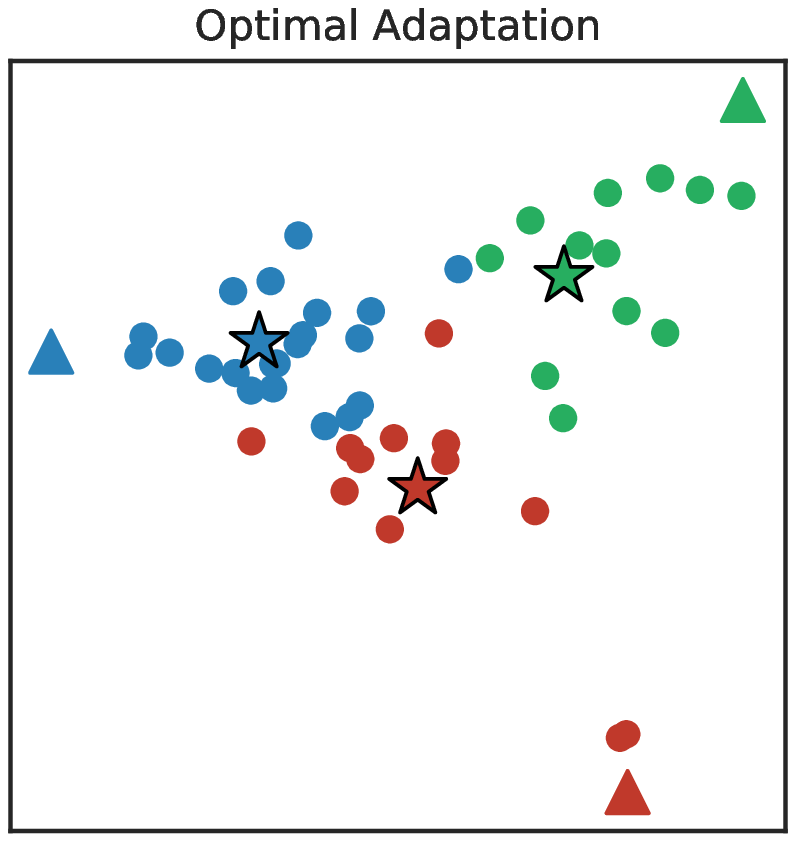}} &
\raisebox{-.5\height}{\includegraphics[width=0.185\textwidth, trim={35 25 28 34},clip]{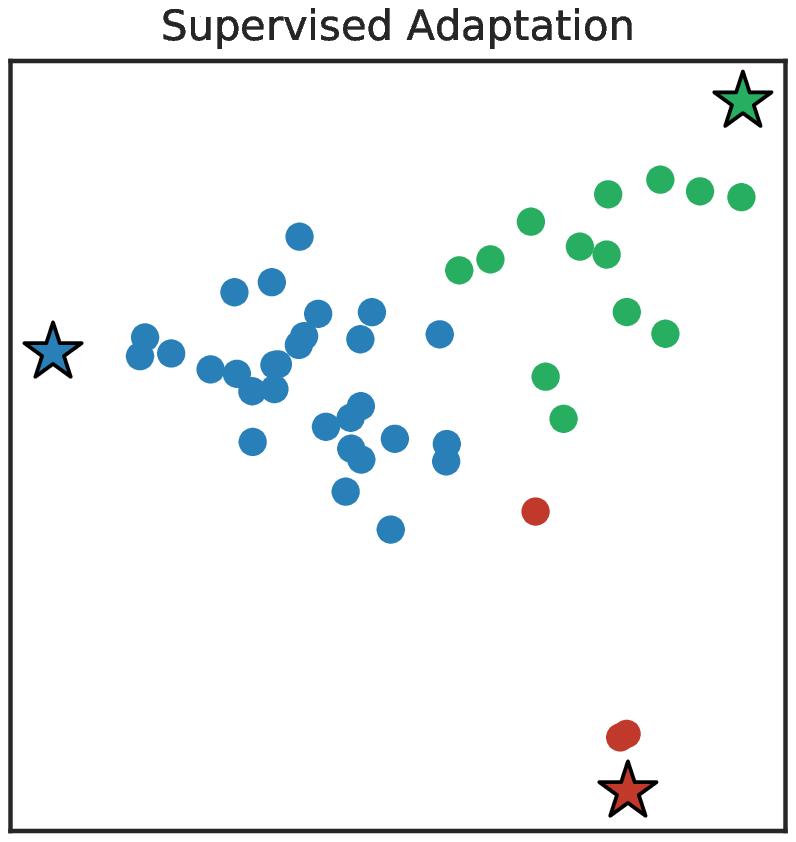}} &
\raisebox{-.5\height}{\includegraphics[width=0.185\textwidth, trim={35 25 28 34},clip]{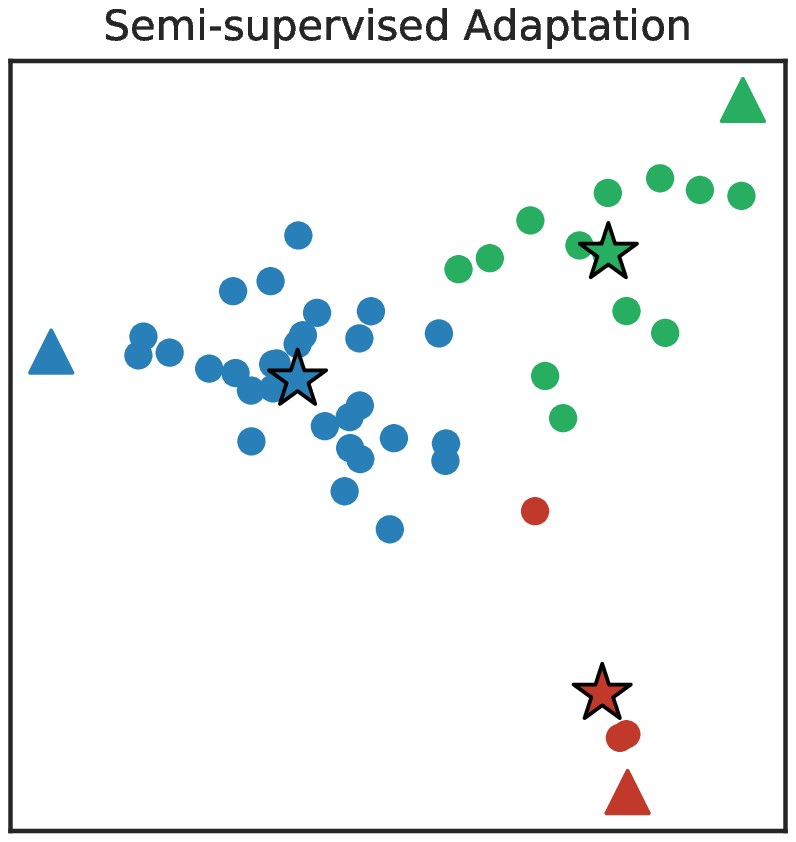}} &
\raisebox{-.5\height}{\includegraphics[width=0.185\textwidth, trim={35 25 28 34},clip]{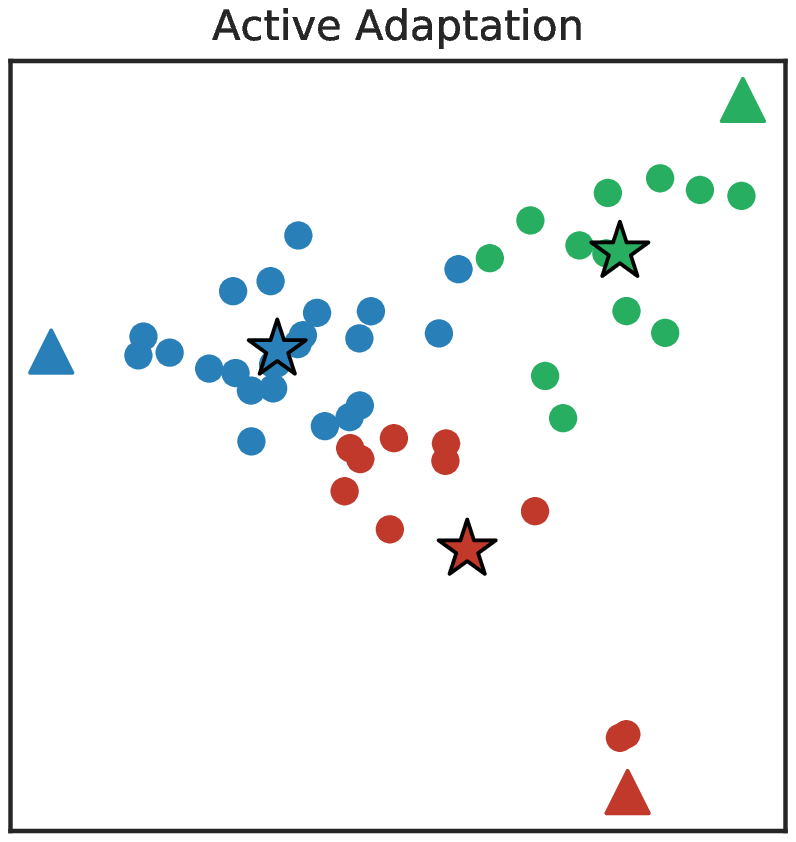}} \\[1.3cm]

(d) &
\raisebox{-.5\height}{\includegraphics[width=0.185\textwidth, trim={35 25 28 34},clip]{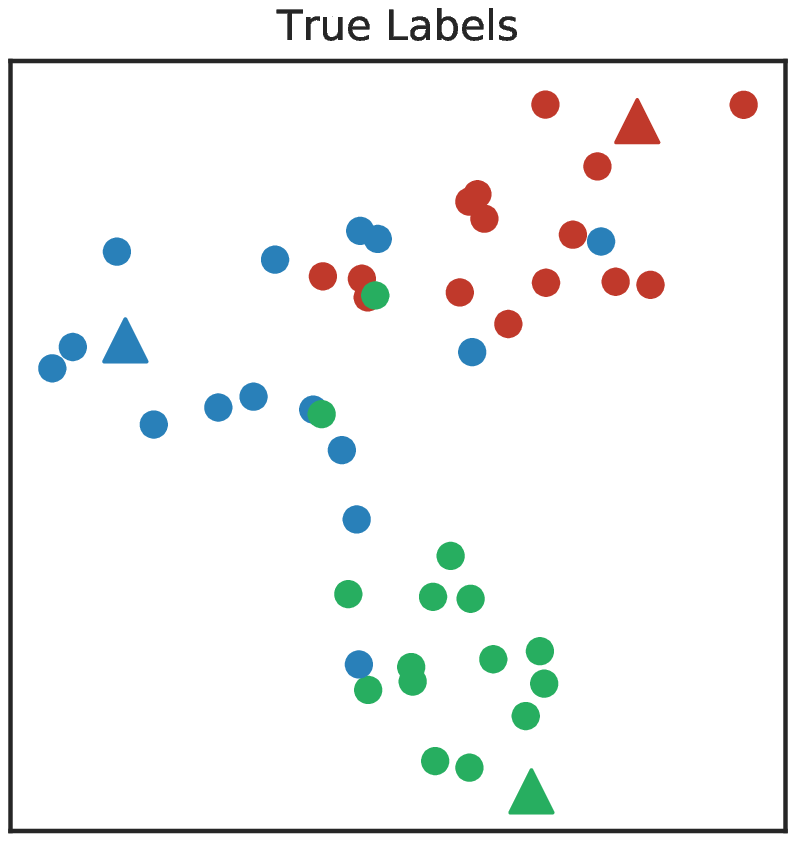}} &
\raisebox{-.5\height}{\includegraphics[width=0.185\textwidth, trim={35 25 28 34},clip]{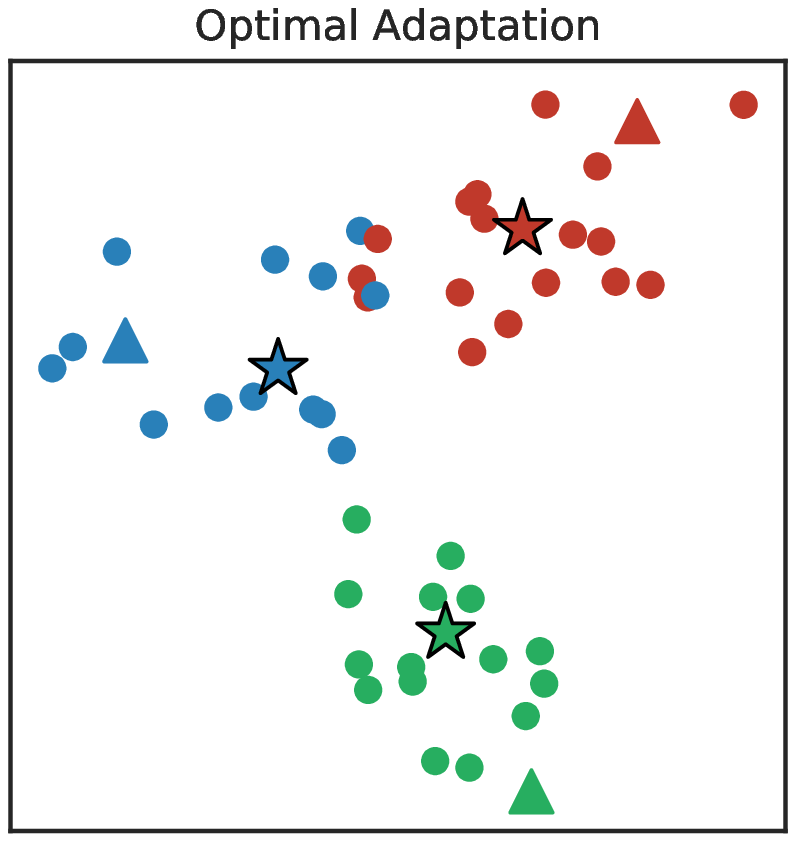}} &
\raisebox{-.5\height}{\includegraphics[width=0.185\textwidth, trim={35 25 28 34},clip]{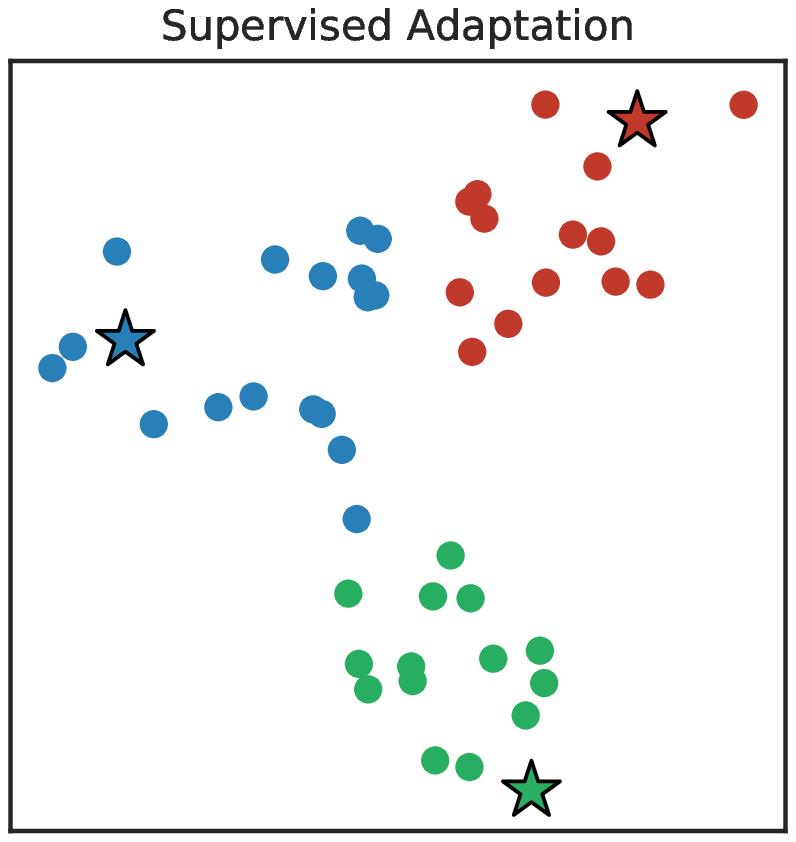}} &
\raisebox{-.5\height}{\includegraphics[width=0.185\textwidth, trim={35 25 28 34},clip]{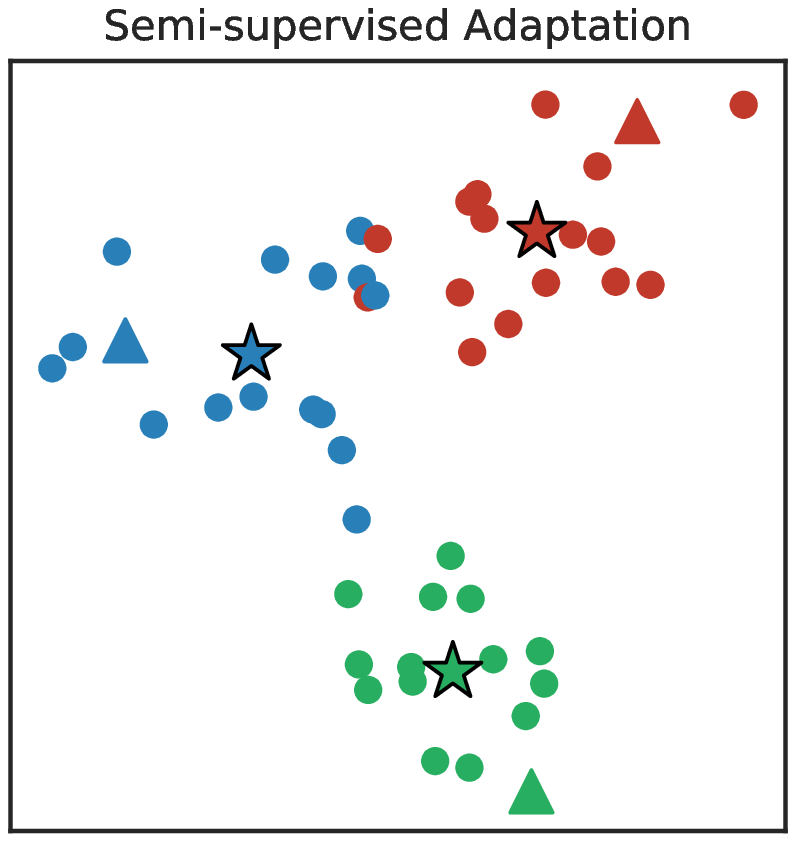}} &
\raisebox{-.5\height}{\includegraphics[width=0.185\textwidth, trim={35 25 28 34},clip]{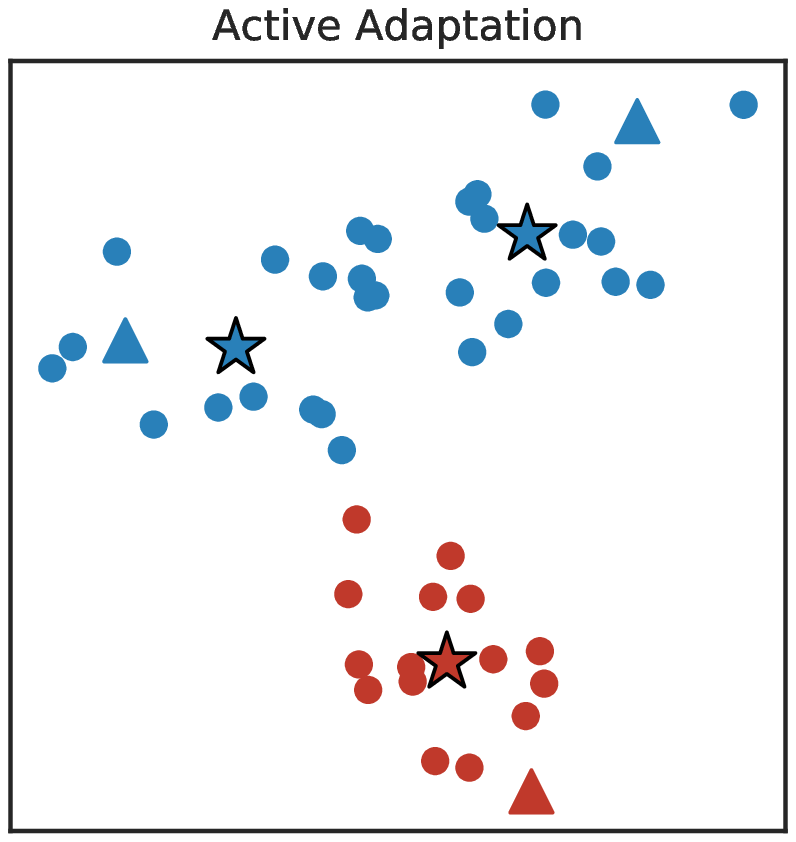}} \\[1.3cm]

(e) &
\raisebox{-.5\height}{\includegraphics[width=0.185\textwidth, trim={35 25 28 34},clip]{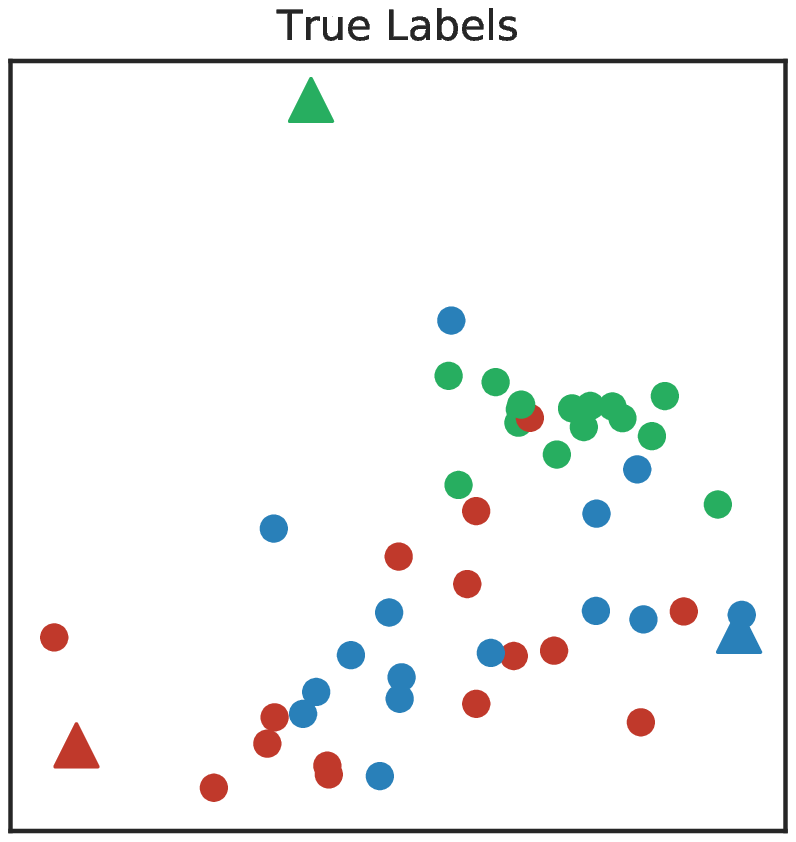}} &
\raisebox{-.5\height}{\includegraphics[width=0.185\textwidth, trim={35 25 28 34},clip]{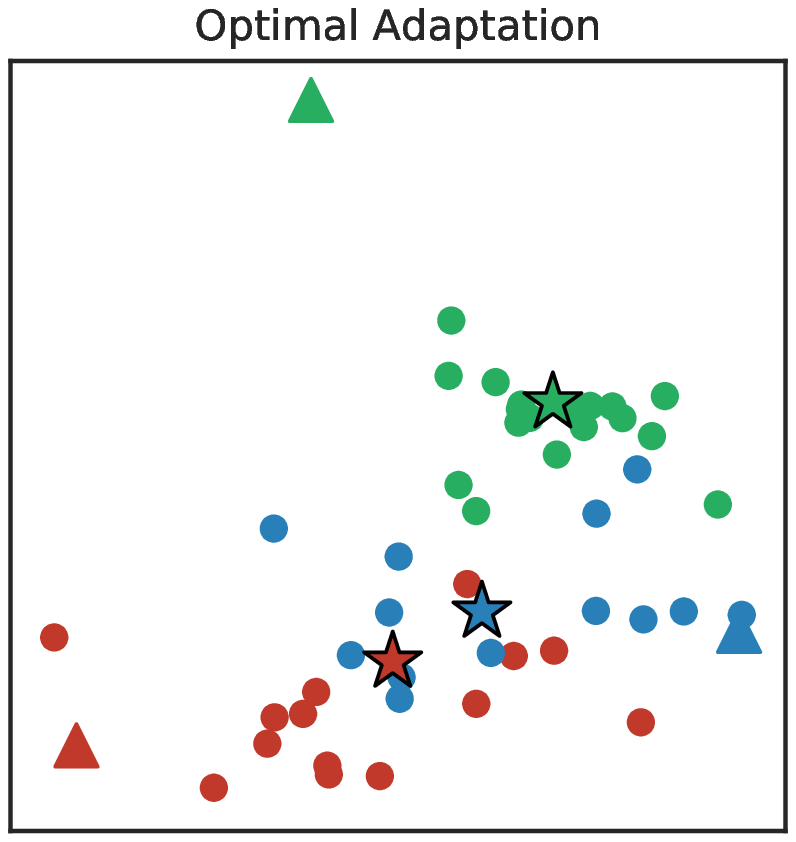}} &
\raisebox{-.5\height}{\includegraphics[width=0.185\textwidth, trim={35 25 28 34},clip]{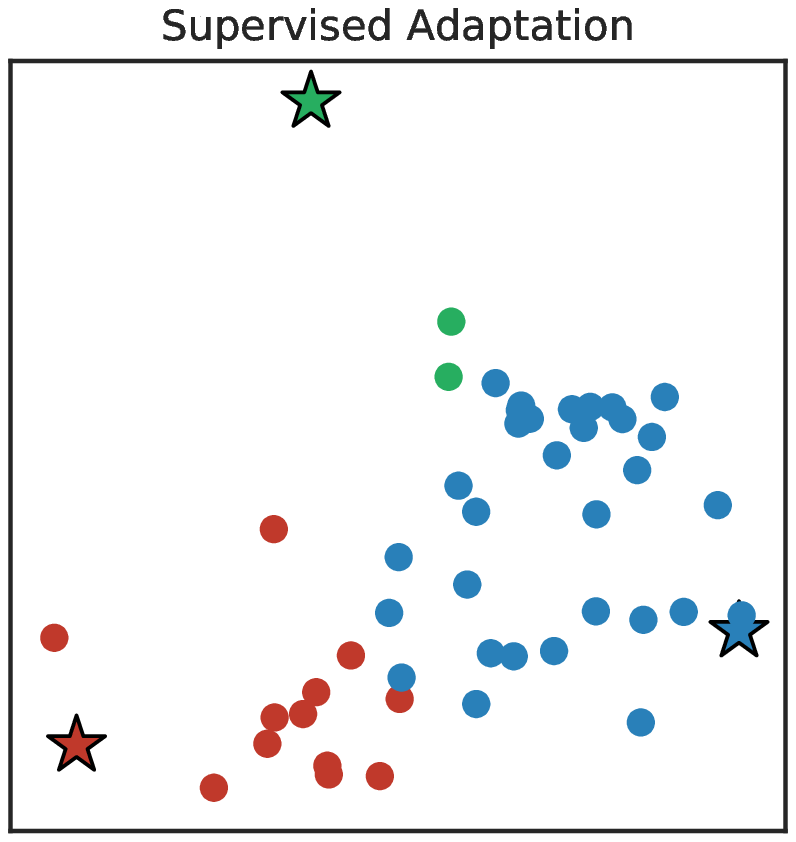}} &
\raisebox{-.5\height}{\includegraphics[width=0.185\textwidth, trim={35 25 28 34},clip]{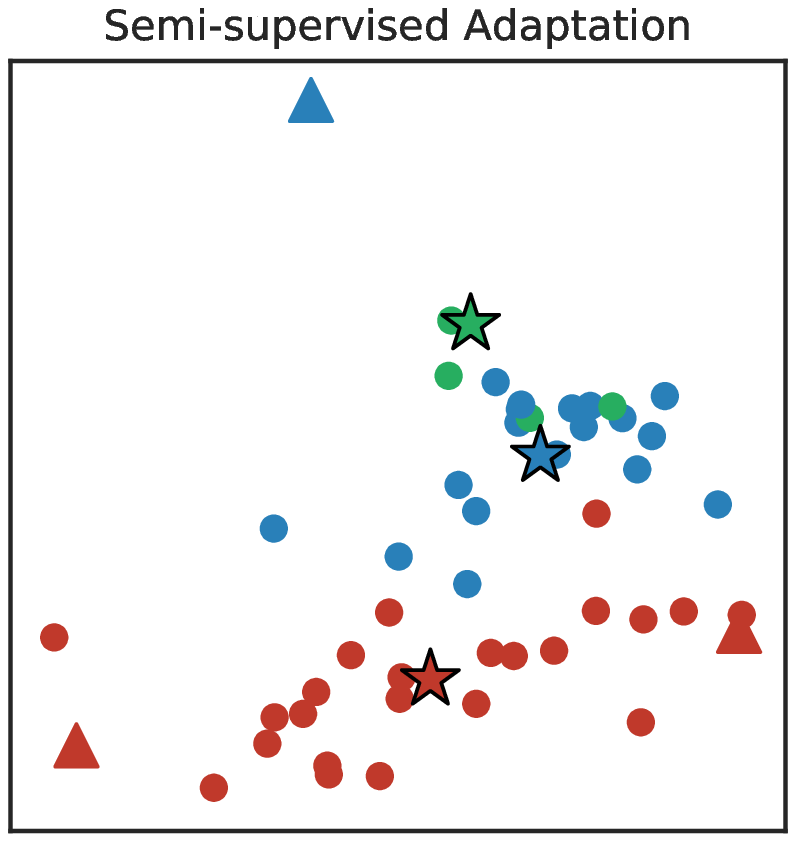}} &
\raisebox{-.5\height}{\includegraphics[width=0.185\textwidth, trim={35 25 28 34},clip]{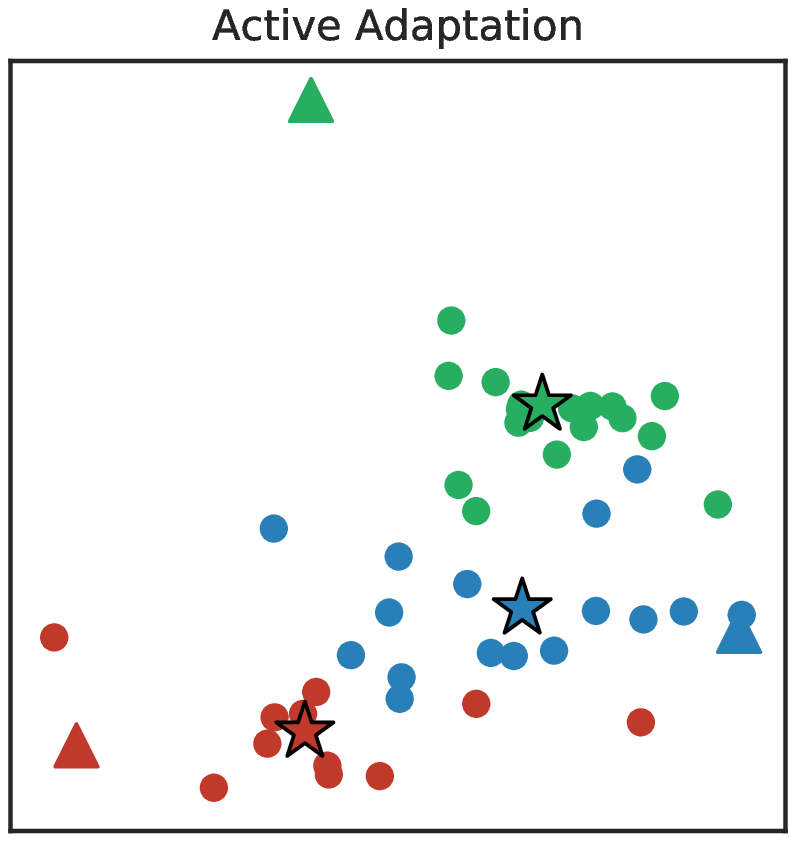}} \\
\end{tabular}
\caption{Illustration of 3-way 1-shot miniImagenet tasks (best viewed in color).
The two-dimensional visualizations are obtained by projecting the 384-dimensional features
onto the principal subspace of the prototypes computed from the support set. The support set is represented using triangle markers, the query set
using circle markers and the prototypes using star markers. The colors represent the true labels in the first column and the labels produced with different adaptation strategies in columns 2--5. Each row corresponds to one 3-way 1-shot task from miniImagenet. 
\textbf{True Labels}: The true labels.
\textbf{Optimal}: Predictions based on the prototype computed from the true labels of all the samples.
\textbf{Supervised}: Predictions based on the prototypes computed from the support set.
\textbf{Semi-supervised}: Predictions based on the prototypes (or cluster means) computed after $K$-means clustering seeded
by the prototypes of the support set.
\textbf{Active}: Predictions based on the prototypes (or cluster means) computed by $K$-means and labeling the clusters using the \textit{Nearest}
approach.
(a)--(b): Tasks are reasonably well clustered and supervised adaptation performs quite well. The labeling is further slightly improved by semi-supervised and active adaptation.
(c): The support sample of the red class is an outlier and using it as a prototype leads to misclassifications.
Even seeding the clustering with these prototypes (semi-supervised classification) leads to incorrect clustering.
However, the active adaptation is able to find a reasonable solution.
(d):~A~failure case of the \textit{Nearest} approach of the active adaptation where the samples are properly clustered but incorrectly labeled.
(e): Samples are not properly clustered in the feature space. The supervised and semi-supervised approaches fail badly, while the active
approach produces somewhat usable results.
}
\label{f:clustering}
\end{figure}

In the experiments with miniImagenet we use a PN trained in the episodic mode as the feature extractor.
We simulate active learning on test tasks by first doing $K$-means clustering in the PN embedding space
and then requesting one labeled example for each cluster using the acquisition functions described earlier.
Note that multiple clusters can be labeled to the same class if the requested labels guide it that way. This is the largest source
of error (see also Fig.~\ref{f:clustering}).
Table~\ref{t:miniimagenet_active_nsamp} presents the classification performance of each strategy for test
tasks with one labeled sample and a varying number of unlabeled samples. There, we also present the accuracy of the oracle clustering.
Overall, the \textsc{margin} approach worked best in our experiments. The 1-shot classification accuracy with 120 unlabeled samples per
class even surpassed the 5-shot accuracy of some well-recognized previous methods. Similar to the semi-supervised scenario,
the four-block architecture scales well with increasing the number of unlabeled samples closely matching the performance of the ResNet in the
case of 120 unlabeled samples per class and even outperforming it while using the \textsc{margin} strategy.

\begin{table}[t]
\caption{Average 1-shot classification accuracy (with 95\% confidence intervals) on miniImagenet for the active learning scenario for different number of unlabeled samples per class ($M$) available at test time.}
\label{t:miniimagenet_active_nsamp}
\vskip 0.15in
\begin{center}
%\begin{small}
\resizebox{\textwidth}{!}{\begin{tabular}{ll|lllll}
\toprule
 & $M$ & Random & Nearest & Entropy & Margin & Oracle \\
\midrule
\multirow{4}{*}{PN (ours)} & 15 & $49.19 \pm 0.88$ & $54.42 \pm 0.34$ & $53.95 \pm 0.38$ & $56.12 \pm 0.39$ & $58.96 \pm 0.70$ \\
& 30 & $49.23 \pm 0.93$ & $54.73 \pm 0.80$ & $56.02 \pm 0.38$ & $57.58 \pm 0.38$ & $60.27 \pm 0.69$ \\
& 60 & $50.73 \pm 0.94$ & $56.12 \pm 0.87$ & $57.63 \pm 0.39$ & $59.24 \pm 0.39$ & $62.09 \pm 0.64$ \\
& 120 & $50.74 \pm 0.94$ & $57.45 \pm 0.86$ & $57.88 \pm 0.39$ & $61.42 \pm 0.37$ & $63.23 \pm 0.65$ \\
\midrule
\multirow{4}{*}{Resnet PN} & 15 & $51.10 \pm 0.48$ & $57.50 \pm 0.46$ & $58.24 \pm 0.39$ & $60.00 \pm 0.40$ & $62.44 \pm 0.34$ \\
& 30 & $51.16 \pm 0.46$ & $57.63 \pm 0.44$ & $59.45 \pm 0.38$ & $60.29 \pm 0.37$ & $62.94 \pm 0.32$ \\
& 60 & $51.36 \pm 0.48$ & $57.68\pm 0.46$ & $59.77 \pm 0.48$ & $60.43 \pm 0.40$ & $63.21 \pm 0.37$ \\
& 120 & $51.56 \pm 0.46$ & $58.09 \pm 0.42$ & $60.19 \pm 0.41$ & $60.49 \pm 0.37$ & $63.71 \pm 0.36$ \\
\bottomrule
\end{tabular}}
%\end{small}
\end{center}
\end{table}

\addtolength{\tabcolsep}{-4.4pt}
\addtolength{\tabcolsep}{4.4pt}

\section{Discussion}
\label{sec:discussion}

In this paper, we extended Prototypical Networks to adapt to new classification tasks in the semi-supervised few-shot learning scenario
when a few labeled examples are accompanied with many unlabeled examples from the same classes. We proposed to use the clustering
approach to semi-supervised classification when the clustering process is guided by the labeled examples. This is different to
recent deep learning papers \citep{rasmus2015semi,miyato2015distributional,laine2016temporal,tarvainen2017weight}
which constrain the classifier using unlabeled data. We also advocated that in many real-world applications it can be possible to request the few labeled examples from the user, which can yield better performance.

The proposed solution of semi-supervised few-shot adaptation is based on doing $K$-means clustering in the embedding space found by Prototypical
Networks. These two methods make a good fit because they make similar assumptions about the data distribution:
In Prototypical Networks, the distribution of each class is represented by its mean and the variances of class distributions are assumed equal.
The same assumptions are made by $K$-means.

The fundamental bottleneck of the proposed approach in improving the classification performance is the ability of the feature extractor
to \emph{cluster unseen data}. Although we used an embedding network trained using Prototypical Networks, the adaptation mechanisms proposed
in this paper can be performed using other feature extractors as well.
A feature extractor explicitly trained to cluster data
can further improve the few-shot classification performance and this is an area of active research \citep{song2016deep,song2017deep,law2017deep}.
Building feature extractors that allow better generalization is largely an unsolved problem and it requires further exploration
\citep[see, e.g.,][]{sabour2017dynamic,hinton2018matrix}. 

\subsubsection*{Acknowledgments}

We would like to thank our colleagues from The Curious AI Company for fruitful discussions.

\bibliography{bibliography.bib}

\begin{thebibliography}{}

\bibitem[Basu {\em et~al.}(2002)Basu, Banerjee, and Mooney]{basu2002semi}
Basu, S., Banerjee, A., and Mooney, R. (2002).
\newblock Semi-supervised clustering by seeding.
\newblock In {\em In Proceedings of 19th International Conference on Machine
  Learning (ICML-2002\/}. Citeseer.

\bibitem[Chapelle {\em et~al.}(2006)Chapelle, Scholkopf, and
  Zien]{chapelle2006semi}
Chapelle, O., Scholkopf, B., and Zien, A., editors (2006).
\newblock {\em Semi-supervised learning\/}.

\bibitem[Cohn {\em et~al.}(2003)Cohn, Caruana, and McCallum]{cohn2003semi}
Cohn, D., Caruana, R., and McCallum, A. (2003).
\newblock Semi-supervised clustering with user feedback.
\newblock {\em Constrained Clustering: Advances in Algorithms, Theory, and
  Applications\/}, {\bf 4}(1), 17--32.

\bibitem[Finn {\em et~al.}(2017a)Finn, Abbeel, and Levine]{finn2017model}
Finn, C., Abbeel, P., and Levine, S. (2017a).
\newblock Model-agnostic meta-learning for fast adaptation of deep networks.
\newblock {\em arXiv preprint arXiv:1703.03400\/}.

\bibitem[Finn {\em et~al.}(2017b)Finn, Yu, Zhang, Abbeel, and
  Levine]{finn2017one}
Finn, C., Yu, T., Zhang, T., Abbeel, P., and Levine, S. (2017b).
\newblock One-shot visual imitation learning via meta-learning.
\newblock {\em arXiv preprint arXiv:1709.04905\/}.

\bibitem[Hinton {\em et~al.}(2018)Hinton, Sabour, and Frosst]{hinton2018matrix}
Hinton, G.~E., Sabour, S., and Frosst, N. (2018).
\newblock Matrix capsules with {EM} routing.
\newblock {\em International Conference on Learning Representations\/}.

\bibitem[Kingma and Ba(2014)Kingma and Ba]{kingma2014adam}
Kingma, D. and Ba, J. (2014).
\newblock Adam: A method for stochastic optimization.
\newblock {\em arXiv preprint arXiv:1412.6980\/}.

\bibitem[Koch {\em et~al.}(2015)Koch, Zemel, and
  Salakhutdinov]{koch2015siamese}
Koch, G., Zemel, R., and Salakhutdinov, R. (2015).
\newblock Siamese neural networks for one-shot image recognition.
\newblock In {\em ICML Deep Learning Workshop\/}, volume~2.

\bibitem[Laine and Aila(2016)Laine and Aila]{laine2016temporal}
Laine, S. and Aila, T. (2016).
\newblock Temporal ensembling for semi-supervised learning.
\newblock {\em arXiv preprint arXiv:1610.02242\/}.

\bibitem[Lake {\em et~al.}(2015)Lake, Salakhutdinov, and
  Tenenbaum]{lake2015human}
Lake, B.~M., Salakhutdinov, R., and Tenenbaum, J.~B. (2015).
\newblock Human-level concept learning through probabilistic program induction.
\newblock {\em Science\/}, {\bf 350}(6266), 1332--1338.

\bibitem[Law {\em et~al.}(2017)Law, Urtasun, and Zemel]{law2017deep}
Law, M.~T., Urtasun, R., and Zemel, R.~S. (2017).
\newblock Deep spectral clustering learning.
\newblock In {\em International Conference on Machine Learning\/}, pages
  1985--1994.

\bibitem[Li {\em et~al.}(2017)Li, Zhou, Chen, and Li]{li2017meta}
Li, Z., Zhou, F., Chen, F., and Li, H. (2017).
\newblock Meta-{SGD}: Learning to learn quickly for few shot learning.
\newblock {\em arXiv preprint arXiv:1707.09835\/}.

\bibitem[Lloyd(1982)Lloyd]{lloyd1982least}
Lloyd, S. (1982).
\newblock Least squares quantization in {PCM}.
\newblock {\em IEEE Transactions on Information Theory\/}, {\bf 28}(2),
  129--137.

\bibitem[Miyato {\em et~al.}(2015)Miyato, Maeda, Koyama, Nakae, and
  Ishii]{miyato2015distributional}
Miyato, T., Maeda, S.-i., Koyama, M., Nakae, K., and Ishii, S. (2015).
\newblock Distributional smoothing with virtual adversarial training.
\newblock {\em arXiv preprint arXiv:1507.00677\/}.

\bibitem[Munkhdalai and Yu(2017)Munkhdalai and Yu]{munkhdalai2017meta}
Munkhdalai, T. and Yu, H. (2017).
\newblock Meta networks.
\newblock {\em arXiv preprint arXiv:1703.00837\/}.

\bibitem[Rasmus {\em et~al.}(2015)Rasmus, Berglund, Honkala, Valpola, and
  Raiko]{rasmus2015semi}
Rasmus, A., Berglund, M., Honkala, M., Valpola, H., and Raiko, T. (2015).
\newblock Semi-supervised learning with {L}adder networks.
\newblock In {\em Advances in Neural Information Processing Systems\/}, pages
  3546--3554.

\bibitem[Ravi and Larochelle(2016)Ravi and Larochelle]{ravi2016optimization}
Ravi, S. and Larochelle, H. (2016).
\newblock Optimization as a model for few-shot learning.

\bibitem[Ren {\em et~al.}(2018)Ren, Triantafillou, Ravi, Snell, Swersky,
  Tenenbaum, Larochelle, and Zemel]{ren1meta}
Ren, M., Triantafillou, E., Ravi, S., Snell, J., Swersky, K., Tenenbaum, J.~B.,
  Larochelle, H., and Zemel, R.~S. (2018).
\newblock Meta-learning for semi-supervised few-shot classification.
\newblock In {\em International Conference on Learning Representations\/}.

\bibitem[Sabour {\em et~al.}(2017)Sabour, Frosst, and
  Hinton]{sabour2017dynamic}
Sabour, S., Frosst, N., and Hinton, G.~E. (2017).
\newblock Dynamic routing between capsules.
\newblock In {\em Advances in Neural Information Processing Systems\/}, pages
  3859--3869.

\bibitem[Scheffer {\em et~al.}(2001)Scheffer, Decomain, and
  Wrobel]{scheffer2001active}
Scheffer, T., Decomain, C., and Wrobel, S. (2001).
\newblock Active hidden {M}arkov models for information extraction.
\newblock In {\em International Symposium on Intelligent Data Analysis\/},
  pages 309--318. Springer.

\bibitem[Shyam {\em et~al.}(2017)Shyam, Gupta, and
  Dukkipati]{shyam2017attentive}
Shyam, P., Gupta, S., and Dukkipati, A. (2017).
\newblock Attentive recurrent comparators.
\newblock {\em arXiv preprint arXiv:1703.00767\/}.

\bibitem[Snell {\em et~al.}(2017)Snell, Swersky, and
  Zemel]{snell2017prototypical}
Snell, J., Swersky, K., and Zemel, R.~S. (2017).
\newblock Prototypical networks for few-shot learning.
\newblock {\em arXiv preprint arXiv:1703.05175\/}.

\bibitem[Song {\em et~al.}(2016)Song, Xiang, Jegelka, and
  Savarese]{song2016deep}
Song, H.~O., Xiang, Y., Jegelka, S., and Savarese, S. (2016).
\newblock Deep metric learning via lifted structured feature embedding.
\newblock In {\em Computer Vision and Pattern Recognition (CVPR), 2016 IEEE
  Conference on\/}, pages 4004--4012. IEEE.

\bibitem[Song {\em et~al.}(2017)Song, Jegelka, Rathod, and
  Murphy]{song2017deep}
Song, H.~O., Jegelka, S., Rathod, V., and Murphy, K. (2017).
\newblock Deep metric learning via facility location.
\newblock In {\em Computer Vision and Pattern Recognition (CVPR)\/}.

\bibitem[Tarvainen and Valpola(2017)Tarvainen and Valpola]{tarvainen2017weight}
Tarvainen, A. and Valpola, H. (2017).
\newblock Weight-averaged consistency targets improve semi-supervised deep
  learning results.
\newblock {\em arXiv preprint arXiv:1703.01780\/}.

\bibitem[Vinyals {\em et~al.}(2016)Vinyals, Blundell, Lillicrap, Wierstra, {\em
  et~al.}]{vinyals2016matching}
Vinyals, O., Blundell, C., Lillicrap, T., Wierstra, D., {\em et~al.} (2016).
\newblock Matching networks for one shot learning.
\newblock In {\em Advances in Neural Information Processing Systems\/}, pages
  3630--3638.

\bibitem[Wagstaff {\em et~al.}(2001)Wagstaff, Cardie, Rogers, Schr{\"o}dl, {\em
  et~al.}]{wagstaff2001constrained}
Wagstaff, K., Cardie, C., Rogers, S., Schr{\"o}dl, S., {\em et~al.} (2001).
\newblock Constrained k-means clustering with background knowledge.
\newblock In {\em ICML\/}, volume~1, pages 577--584.

\bibitem[Zagoruyko and Komodakis(2016)Zagoruyko and
  Komodakis]{zagoruyko2016wide}
Zagoruyko, S. and Komodakis, N. (2016).
\newblock Wide residual networks.
\newblock {\em arXiv preprint arXiv:1605.07146\/}.

\end{thebibliography}

\clearpage

\appendix

\section{Sensitivity to the Number of Shots}

\citet{snell2017prototypical} showed that the classification performance of Prototypical Networks is sensitive to the number $N$ of
classes per task during training and that it was necessary to match the number $k$ of samples per class during training and testing.
We believe that using a larger number of classes ($N$ way) works as a regularizer. However, varying $N$ effectively changes the batch
size and therefore the learning rate needs to be adapted to $N$. By tuning the learning rate for different $N$,
we observed a smaller effect of this regularization on the adaptation accuracy.

The dependency on $k$ implies that the embedding learned by the network does not generalize well to a different
number of samples during test time. This is inconvenient especially if the number of the labeled example is unknown in advance and
it can grow, for example, as a result of interaction with the user.
 To address this problem, we propose to use a varying number of samples per class
during training for better generalization to the number of shots during test time.
The results in Table~\ref{t:k-shot} illustrate that this strategy is effective and reduces the sensitivity to $k$ during training.

\begin{table}[h]
\caption{Results of fully supervised adaptation using PN as a function
of the number $k$ of samples per class during training. $[1 .. 5]$ denotes varying
$k$ in the range $[1, \: 5]$.}
\label{t:k-shot}
\begin{center}
\begin{tabular}{lc|ll}
\toprule
\multicolumn{2}{c|}{\multirow{2}{*}{Training shot}} & \multicolumn{2}{c}{Testing shot} \\
& & 1-shot & 5-shot \\
\midrule
\multirow{3}{*}{PN (ours)} & 1 & $48.06 \pm 0.82$ & $63.20 \pm 0.77$ \\
& 5 & $42.71 \pm 0.75$ & $64.65 \pm 0.72$ \\
& $[1 .. 5]$ & $48.12 \pm 0.78 $ & $64.84 \pm 0.70$ \\
\midrule
\multirow{3}{*}{Resnet PN} & 1 & $51.69 \pm 0.47$ & $67.43 \pm 0.63$ \\
& 5 & $50.44 \pm 0.48$ & $69.57 \pm 0.59$ \\
& $[1 .. 5]$ & $51.40 \pm 0.46$ & $68.97 \pm 0.55$ \\
\bottomrule
\end{tabular}
\end{center}
\end{table}

\end{document}